%% file: main.tex
\documentclass[sigconf]{acmart}

\usepackage{multirow}
\usepackage{subcaption}
\usepackage{graphicx}
\usepackage{algorithm}
\usepackage{algpseudocode}
\usepackage[dvipsnames]{xcolor}

\algrenewcommand\algorithmicrequire{\textbf{Input:}}
\algrenewcommand\algorithmicensure{\textbf{Output:}}

\AtBeginDocument{%
  }

\setcopyright{acmlicensed}
\copyrightyear{2018}
\acmYear{2018}
\acmDOI{XXXXXXX.XXXXXXX}
\acmConference[Conference acronym 'XX]{Make sure to enter the correct
  conference title from your rights confirmation email}{June 03--05,
  2018}{Woodstock, NY}
\acmISBN{978-1-4503-XXXX-X/2018/06}

\begin{document}

\title{Think with Grounding: Curriculum Reinforced Reasoning with Video Grounding for Long Video Understanding}

\author{Houlun Chen}
\affiliation{%
  \institution{Tsinghua University}
  \city{Beijing}
  \country{China}}
\email{}

\author{Xin Wang}
\affiliation{%
  \institution{Tsinghua University}
  \city{Beijing}
  \country{China}
}

\author{Guangyao Li}
\affiliation{%
  \institution{Tsinghua University}
  \city{Beijing}
  \country{China}
}

\author{Yuwei Zhou}
\affiliation{%
  \institution{Tsinghua University}
  \city{Beijing}
  \country{China}
}

\author{Yihan Chen}
\affiliation{%
  \institution{Tsinghua University}
  \city{Beijing}
  \country{China}
}

\author{Jia Jia}
\affiliation{%
  \institution{Tsinghua University}
  \city{Beijing}
  \country{China}
}

\author{Wenwu Zhu}
\affiliation{%
  \institution{Tsinghua University}
  \city{Beijing}
  \country{China}
}
\renewcommand{\shortauthors}{Chen et al.}

\begin{abstract}

Long video understanding~(LVU) is challenging due to rich and complicated multimodal clues in long temporal range. Current methods adopt reasoning to improve the model's ability to analyze complex video clues in long videos via text-form reasoning. However, the existing literature suffers from the fact that the text-only reasoning under fixed video context may exacerbate hallucinations since detailed crucial clues are often ignored under limited video context length due to the temporal redundancy of long videos.
To address this gap, we propose \textbf{Video-TwG}, a curriculum reinforced framework that employs a novel \textit{\textbf{T}hink-\textbf{w}ith-\textbf{G}rounding} paradigm, enabling video LLMs to {actively} decide when to perform on-demand grounding during interleaved text–video reasoning, selectively zooming into question-relevant clips only when necessary. Video-TwG can be trained end-to-end in a straightforward manner, without relying on complex auxiliary modules or heavily annotated reasoning traces.
In detail, we design a Two-stage Reinforced Curriculum Strategy, where the model first learns \textit{think-with-grounding} behavior on a small short-video GQA dataset with grounding labels, and then scales to diverse general QA data with videos of diverse domains to encourage generalization. Further, to handle complex \textit{think-with-grounding} reasoning for various kinds of data, we propose TwG-GRPO algorithm which features the fine-grained grounding reward, self-confirmed pseudo reward and accuracy-gated mechanism.  
Finally, we propose to construct a new TwG-51K dataset that facilitates training.
Experiments on Video-MME, LongVideoBench, and MLVU show that Video-TwG consistently outperforms strong LVU baselines. Further ablation validates the necessity of our Two-stage Reinforced Curriculum Strategy and shows our TwG-GRPO better leverages diverse unlabeled data to improve grounding quality and reduce redundant groundings without sacrificing QA performance.

\end{abstract}







\setcopyright{none}
\settopmatter{printacmref=false}
\renewcommand\footnotetextcopyrightpermission[1]{}
\pagestyle{plain}

\maketitle

\input{my_intro}

\input{my_related_works}

\input{my_method}

\input{my_experiments}

\input{my_conclude}

% \begin{acks}
% \end{acks}

\bibliographystyle{ACM-Reference-Format}
\bibliography{main}

\appendix

\end{document}

%% file: my_intro.tex
\section{Introduction}

Long video understanding (LVU)~\cite{wu2021towards,fu2025video,wu2024longvideobench,zhou2025mlvu,wang2025lvbench} has become a cornerstone for real-world multimodal information access. Users increasingly pose natural-language questions about movies, vlogs, news, and instructional videos, expecting accurate answers grounded in long-form video content. Recent benchmarks further push LVU toward realistic, hour-scale videos and compositional reasoning, exposing substantial gaps in current video large language models (LLMs) with respect to long-horizon perception and reasoning~\cite{fu2025video,wu2024longvideobench,zhou2025mlvu}.

Many works address LVU from multiple perspective and achieve good performance, such as long context modeling~\cite{chen2024longvila,shen2025long,ren2025vamba,shu2025video}, memory-centric designs~\cite{song2024moviechat,santos2025infty,Diko_2025_CVPR}, token compression~\cite{zhang2025llava,shen2024longvu,meng2024deepstack,Tao_2025_CVPR,Wang_2025_ICCV}, adaptive sampling~\cite{Tang_2025_CVPR,Liu_2025_CVPR}, video agent pipelines~\cite{wang2025videotree,ren2025videorag,luo2024video}, \textit{e.t.c}.
Since long videos contain rich and complicated multimodal clues in long temporal range, recent video reasoning LLMs~\cite{feng2025video,li2025videochat,chen2025scaling} benefit from test-time-scaling via conducting text-form reasoning to better analyze the complex video clues, which contributes to LVU significantly. However, they conduct text-only reasoning under fixed video context, suffering from hallucinations since the important detailed information might be ignored under limited video context length due to the temporal redundancy of long videos, where only a few sparsely distributed moments are relevant for answering a given question, and thus, achieve sub-optimal results.

\begin{figure}[h]
    \centering
    \includegraphics[width=\linewidth]{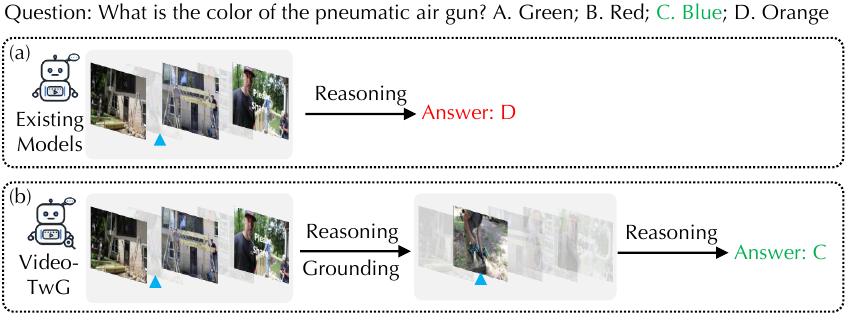}
    \caption{Conceptual level comparison between our Video-TwG and existing video reasoning models. The frame with \textcolor{blue}{blue} triangle means the important detail for this question. Existing reasoning models reach the wrong answer via reasoning on missed video context, while we address it via leveraging grounding during reasoning to dynamically perceive important clues.}
    \Description{}
    \label{fig:intro-concept-level}
\end{figure}

To address this problem, we propose \textbf{Video-TwG}, a curriculum reinforced  framework where we employ a novel \textit{\textbf{T}hink-\textbf{w}ith-\textbf{G}rounding} paradigm, which equips video LLMs with explicit on-demand video grounding behavior during reasoning, without relying on complex auxiliary modules or costly supervised fine-tuning reasoning traces. As shown in Fig.~\ref{fig:intro-concept-level}, instead of passively reasoning from a fixed, statically sampled video context, our method follows the insight of Retrieval-Augmented-Generation~(RAG) and iteratively decides what evidence it needs next and issues grounding actions to zoom into query-relevant video clips with fine-grained representation.

In details, our Video-TwG mainly contains the following parts. To reduce the training difficulty, we propose a Two-stage Reinforced Curriculum Strategy. In the first stage, the model learns \textit{think-with-grounding} paradigm on easier, short-video GQA data. In the second stage, the training scales to general QA data on videos of various domains and duration, where grounding labels are largely unavailable. In this stage, the model is encouraged to generalize its \textit{think-with-grounding} ability to richer scenarios. Furthermore, to handle the complicated reasoning trajectories for various kinds of data, we propose the TwG-GRPO algorithm, where we assign the fine-grained grounding reward for GQA data, and specifically design the self-confirmed pseudo reward for large amounts of general QA data to encourage high-quality grounding actions, and we apply the accuracy-gated mechanism to avoid the optimization conflicts between accuracy and grounding targets.
To support our training, we propose to construct TwG-51K, a novel dataset which combines (a) a small subset of video GQA data with grounding annotations and (b) a large collection of unlabeled general video QA data for generalization purpose. 

Empirically, without relying on heavily annotated reasoning traces for supervised fine-tuning, Video-TwG consistently outperforms strong baselines across three mainstream benchmarks after two-stage curriculum training: Video-MME~\cite{fu2025video}, LongVideoBench~\cite{wu2024longvideobench}, and MLVU~\cite{zhou2025mlvu}. Compared to Qwen2.5-VL-7B~\cite{bai2025qwen2}, our model achieves improvements of  7.0, 5.3, and 7.1 under low-resolution inputs, and 2.5, 3.9, and 5.0 under high-resolution inputs on these benchmarks, respectively.
Ablation study validates the necessity of our Two-stage Reinforced Curriculum Strategy and shows that our TwG-GRPO effectively helps the model to learn \textit{think-with-grounding} from various data, especially in unlabeled data. It's noted with the self-confirmed pseudo reward, the model learns to conduct grounding actions more adaptively, improving the grounding quality and reducing grounding actions by 15.7\% without sacrificing QA performance.

In summary, our contributions are fourfold:
\begin{itemize}
    \item We propose Video-TwG, a curriculum reinforced framework that enables video LLMs to iteratively think with grounding for long video understanding, without heavy external modules or costly supervised reasoning traces.
    \item We introduce a Two-stage Reinforced Curriculum Strategy to reduce the training difficulty and TwG-GRPO, a novel algorithm which combines fine-grained grounding rewards, self-confirmed pseudo rewards, and an accuracy-gated mechanism to jointly improve answer correctness and grounding quality.
    \item We construct a new dataset, TwG-51K, which can facilitate the training through providing both labeled and unlabeled data.
    \item Extensive experiments on Video-MME, LongVideoBench, and MLVU show consistent advantages over strong LVU baselines.
\end{itemize}

%% file: my_related_works.tex
\section{Related Works}

\subsection{Long Video Understanding}

Long video understanding (LVU) is the fundamental cornerstone for real-life applications and is quite challenging for current video LLMs.
Recent LVU methods mainly fall into the following directions.
(i) {Long-context modeling} scales video LLMs to longer inputs via architectural/efficiency improvements~\cite{chen2024longvila,shen2025long,ren2025vamba,shu2025video} with hour-scale attempts.
(ii) {Memory-centric designs} maintain learnable or external memory to accumulate salient cues over time~\cite{song2024moviechat,santos2025infty,Diko_2025_CVPR}.
(iii) {Token compression} reduces the cost of long sequences~\cite{zhang2025llava,shen2024longvu,meng2024deepstack,Tao_2025_CVPR,Wang_2025_ICCV}.
(iv) {Adaptive sampling} selects informative frames/clips under budget~\cite{Tang_2025_CVPR,Liu_2025_CVPR} and highlights "needle-in-a-haystack" temporal localization~\cite{Ye_2025_CVPR}. Meanwhile, realistic benchmarks like MLVU expose persistent length-induced degradation~\cite{zhou2025mlvu}.
(v) {video agent systems} decompose the long video question into several intermediate steps and design cascaded pipelines~\cite{wang2025videotree,ren2025videorag,luo2024video,yang2025vca}
Above all, the insight is to model long-range relations among multimodal cues and allocate computation to the most informative parts of long videos.
This motivates our Video-TwG framework that enables video LLMs to invoke on-demand grounding during reasoning: reasoning interprets complex cues, while grounding pinpoints the most relevant video clips.

\subsection{Video Reasoning}
Recent "test-time scaling" suggest that allocating more computation at inference can further unlock LLM capabilities. Among this, LLM reasoning (e.g., CoT~\cite{wei2022chain}/ToT~\cite{yao2023tree}) has shown clear scaling benefits for solving complex problems, and recent work further demonstrates that post-training with reinforcement learning can substantially strengthen reasoning behaviors~\cite{guo2025deepseek}.
Building on these advances, video reasoning emphasizes producing process-like intermediate traces, like stepwise rationales, to improve output quality and interpretability, rather than only emitting final answers, to better grasp multimodal cues and analyze their relations.
Current post-training reasoning methods, often RL-flavored, encourage models to produce more faithful, verifiable reasoning traces, especially for temporal understanding cases~\cite{feng2025video,Luo_2025_NeurIPS_VER,Huang_2025_NeurIPS_VadR1,Wang_2025_NeurIPS_TimeR1,feng2025video,li2025videochat,chen2025scaling}. 
Besides, new datasets/benchmarks provide reasoning-trace supervision for video understanding and analysis, showing that temporal localization and evidence alignment remain key bottlenecks for current video LLMs~\cite{Nagrani_2025_ICCV,Pei_2025_NeurIPS_EgoThinker}.
However, most works optimize text-form reasoning under fixed video context, which doesn't work well in LVU due to its high temporal redundancy, whereas our Video-TwG triggers dynamic grounding actions during reasoning to retrieve and zoom into most relevant clips, letting the reasoning enhance perception.

\subsection{Video Grounding}
Video grounding localizes the start and end timestamps of a clip that semantically matches a natural-language query~\cite{gao2017tall,anne2017localizing,lan2023survey,chen2023curriculum,huang2025identity,chen2024multi,feng2023llm4vg,chen2024verified}. 
With the rise of Video LLMs, grounding is increasingly used to test whether models can consistently point to temporal evidence rather than only generate plausible descriptions. 
Recent works~\cite{chen2025localizing,Wu_2025_CVPR,Deng_2025_CVPR,Zeng_2025_TimeSuite,Yang_2025_ICCV,Pramanick_2025_ICCV,Zhang_2025_ICCV,Li_2025_UniTime,Jung_2025_CVPR} advance Video-LLM grounding by improving temporal referencing and grounded training/evaluation, yielding better accuracy on standard benchmarks.
Different from the query-only setting, we treat video grounding as an RAG-style retrieval step conditioned on the Video-LLM’s accumulated video-text reasoning context.
At each step, the model first retrieves the most relevant clip given the video and prior reasoning, then re-loads the grounded clip with fine-grained representation into the context to support subsequent reasoning.

%% file: my_method.tex
\section{Video-TwG Framework}

\begin{figure*}[h]
  
    \centering
    \includegraphics[width=\linewidth]{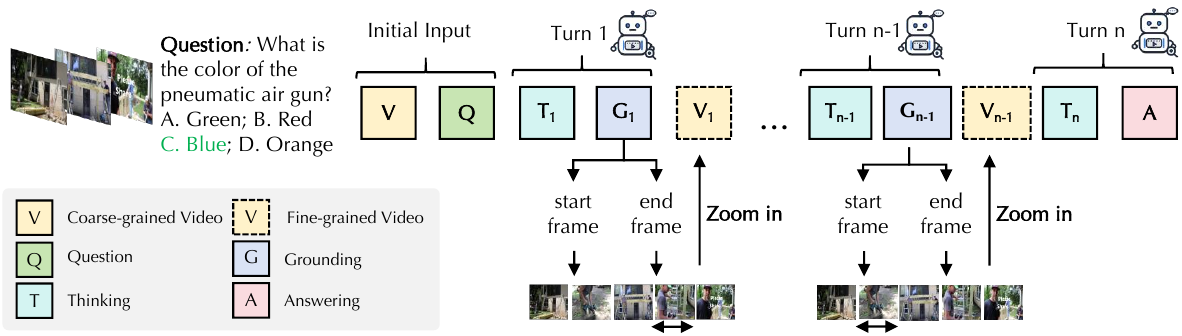}
    \caption{The trajectory of \textit{think-with-grounding}. Initially, the coarse-grained video and question are given and in each turn, the model gives a thinking process and an action based on the history interaction. If the action is grounding, the grounded video clip is zoomed in with a fine-grained representation and added to the context. If the action is answering, the reasoning process stops.}
    \Description{}
    
\end{figure*}

\subsection{Preliminary}
\label{preliminary}

Our framework formulates \textit{think-with-grounding} to multi-turn dialog.
Given a question \(Q\) and a video \(V\), the video LLM \(\mathcal{M}\) is required to give its answer \({a}\) in a reasoning chain consisting of grounding(\(g\)) and answering(\(a\)) actions along with thinking process(\(t\)) in a maximum of \(K\) turns. The \(k\)-th turn is formulated to:
\begin{equation}
    (t_{k},g_{k})\ \text{or}\ (t_{k},{a}) = \mathcal{M}(V,p,t_1,g_1,v_1,p_1,...,t_{k-1},g_{k-1},v_{k-1},p_{k-1}),
\end{equation}
where \(p\) and \(p_1,...,p_{k-1}\) are prompts for initial and intermediate turns containing the question \(Q\) and the task instruction. In each turn, video LLM \(\mathcal{M}\) takes the interaction history as input and outputs the next thinking \(t_{k}\) and action, \textit{i. e.} grounding action \(g_{k}\) or answering action \({a}\). The thinking \(t\), grounding action \(g\) and answering action \({a}\) are in the format of
\begin{equation}
\begin{aligned}
t &\leftarrow \text{\textless think\textgreater thinking process \textless/think\textgreater}; \\
g &\leftarrow \text{\textless ground\textgreater start frame, end frame \textless/ground\textgreater}; \\
a &\leftarrow \text{\textless answer\textgreater final answer \textless/answer\textgreater}. \\
\end{aligned}
\end{equation}
After each turn with the grounding action, the grounded video clip \(v_{k}\) is concatenated to the history.

\subsection{TwG-51K Dataset Construction}

To facilitate training, we construct the TwG-51K dataset, comprising 50,744 multiple-choice video QA samples in total. Among these, 8,195 samples have ground-truth question-relevant video grounding annotations, while 42,549 samples lack such annotations.
(i) For samples with video grounding annotations, we collect Grounded Question Answering~(GQA) data from NExT-GQA~\cite{xiao2024can} and CG-Bench~\cite{chen2024cg}, where the video clip that contains important clues for the answer are annotated.  NExT-GQA provides 10.5K explicit grounding annotations, and we filter out samples whose videos are shorter than 20 seconds and finally obtain 7,123 samples from NExT-GQA. CG-Bench features long videos between 10 and 80 minutes with video grounding annotations, which consists of 12,129 questions from 1,219 videos. We collect videos from CG-Bench-mini and filter out samples where the IoU between the ground-truth video grounding clip and the entire video is smaller than 0.01 to facilitate training, and finally obtain 1,072 samples.
(ii) To scale and diversify training data, we randomly sample a subset of LLaVA-Video-178K~\cite{zhang2024video}, a commonly used multi-domain general video instruction tuning dataset consisting of 1.3M questions from 178K videos, and finally obtain 42,549 multiple-choice training samples.

\subsection{Multi-Grained Video Representation}

Since the initial video is to provide global clues while the dynamically grounded video is to provide detailed information, we adopt multi-grained video representation for different videos.
For the initial video input \(V\), we represent it with \(F_{\text{coarse}}\) frames and \(f_{\text{coarse}}\) tokens per frame in a coarse-grained way, while \(F_\mathrm{fine}\) frames and \(f_\mathrm{fine}\) tokens per frame in a fine-grained way for the dynamically grounded video segment \(v_k\)~(generally, \(F_{\mathrm{coarse}}>F_{\mathrm{fine}}\), \(f_\mathrm{coarse} < f_\mathrm{fine}\)).
To better support videos with various durations and reduce the search scope, in the grounding action, the start and end frames are the frame indexes in the initial video representation, \textit{i.e.} from 0 to \(F_\mathrm{coarse}-1\). When cutting the corresponding video clip, they are converted into real seconds.

\subsection{Two-stage Reinforced Curriculum Strategy}

Since the base video LLM is not capable of benefiting from the \textit{think-with-grounding} stably, to reduce the training difficulty, we design a two-stage curriculum. For the first stage, we train the model with short video GQA data with grounding labels, \textit{i.e.} NExT-GQA part in TwG-51K, serving as cold start, to activate the model to follow the \textit{think-with-grounding} pattern and reach the answer via zooming in the correct video clip. Next, for the second stage, we train the model on the entire TwG-51K dataset, most of which lack grounding labels, to improve the model's generalization via thinking with adaptive grounding.

We skip the SFT and adopt curriculum reinforced learning for both stages, since the model is observed to have the potential to think with grounding, which is demonstrated in subsequent experiments. Benefiting from our RL-only solution, our Video-TwG can be easily improved without heavy SFT annotation works in a lighter way.

\subsection{TwG-GRPO Algorithm}
\begin{figure*}[h]
  
    \centering
    \includegraphics[width=\linewidth]{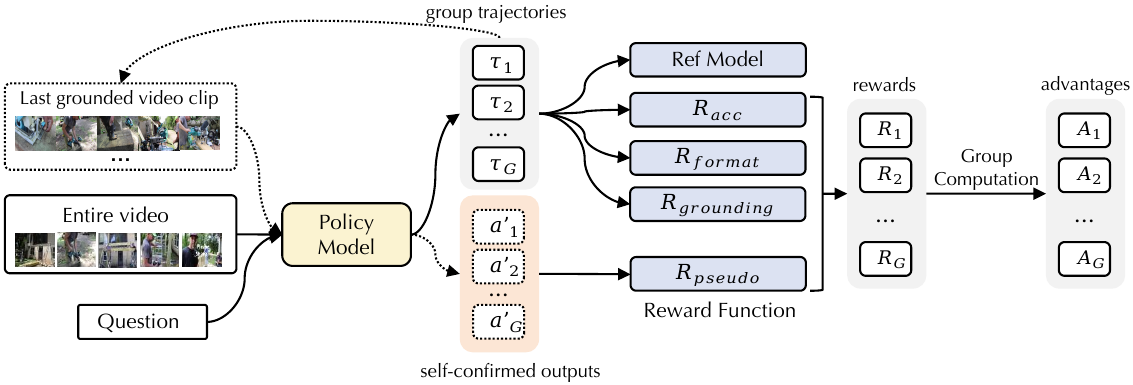}
    \caption{The illustration of our proposed TwG-GRPO algorithm.}
    \label{fig:twg-grpo-framework}
    \Description{}
    
\end{figure*}

As shown in Fig.~\ref{fig:twg-grpo-framework}, we propose a TwG-GRPO algorithm to support RL training in various data, which is based on Group Relative Policy Optimization~(GRPO)~\cite{shao2024deepseekmath}, for the sake of memory-friendly nature. We adopt multi-turn optimization with trajectory-level reward for computation efficiency, where a trajectory \(\tau\) with \(k\) turns follows the format: [\{user: \(V\), \(p\)\},\{assistant: \(t_1\),\(g_1)\) \},\{user: \(v_1\), \(p_1\)\},...,\{assistant: \(t_{k-1}\), \(g_{k-1}\) \},\{user: \(v_{k-1}\), \(p_{k-1}\) \},\{assistant: \(t_{k}\) , \(a\) \}]. For each sample, we collect rollouts and Algorithm~\ref{alg:twg-grpo} shows the rollout process in our multi-turn \textit{think-with-grounding} setting.
After rolling out, we follow the GRPO algorithm to calculate the trajectory advantages in the group:
\begin{equation}
A_{i} = \frac{R_{i} - \mu(\{R_{j}\})}{\sigma( \{R_{j}\} )},
\end{equation}
where the \(\mu(\{R_{j}\})\) and \(\sigma( \{R_{j}\} )\) is the mean and standard deviation of the rewards in each group. The policy video LLM model is optimized by:
\begin{equation}
\begin{aligned}
r_i(\theta) &\triangleq \frac{\mathcal{M}_{\theta}(\tau_i\mid q)}{\mathcal{M}_{\theta_{\mathrm{old}}}(\tau_i\mid q)}, \qquad
\bar r_i(\theta) \triangleq \operatorname{clip}\!\big(r_i(\theta),\,1-\epsilon,\,1+\epsilon\big), \\
\mathcal{J}_{\mathrm{GRPO}}(\theta)
&= \mathbb{E}\!\left[
\frac{1}{G}\sum_{i=1}^{G}
\min\!\big(r_i(\theta)A_i,\;\bar r_i(\theta)A_i\big)
\right]  - \beta\,D_{\mathrm{KL}}\!\left(\mathcal{M}_{\theta}\,\|\,\mathcal{M}_{\mathrm{ref}}\right).
\end{aligned}
\end{equation}
Besides, we deploy dynamic resampling~\cite{yu2025dapo} technique when a group does not provide valid learning signals to improve training stability and accelerate convergence.

Since vanilla accuracy and format rewards can not handle complex \textit{think-with-grounding} reasoning for various data, especially unlabeled general QA data, the rewards are carefully designed, which is elaborated on in the following. We adopt accuracy and format rewards according to convention, and for a trajectory with the grounding action, we calculate a grounding-related reward to give feedback to the grounding action, where we design Fine-grained Grounding Reward and Self-Confirmed Pseudo Reward for the training sample with and without grounding labels, respectively. To avoid the optimization conflicts between accuracy and grounding, we apply the accuracy-gated mechanism to give higher priority to accuracy.

\noindent \textbf{Accuracy and Format Rewards.}
Following ~\cite{guo2025deepseek}, the accuracy and format rewards are included. The accuracy reward evaluates the correctness of the answer in the last turn.
\begin{equation}
R_{\mathrm{acc}} = \begin{cases}
1.0\ \ \text{final answer is correct}\\
0.0\ \ \text{otherwise}.
\end{cases}
\end{equation}
The format reward evaluates if a trajectory follows the grounding or answering action format in all turns.
\begin{equation}
    R_\mathrm{format} = \begin{cases}
0.2\ \ \text{all turns follow the right format}\\
0.0\ \ \text{otherwise}.
\end{cases}
\end{equation}

\noindent \textbf{Fine-grained Grounding Reward.}
For data with video grounding annotations, the \(R_\mathrm{grounding}\) is calculated by
\begin{equation}
\begin{aligned}
    R_\mathrm{soft} &= \text{IoU}({{g_\mathrm{last}}},{g_\mathrm{gt}}) \\
    R_\mathrm{hard} &= 0.5 \cdot \mathbb{I} (\text{IoU}({{g_\mathrm{last}}},{{g_\mathrm{gt}}})>0.0) \\
    R_\mathrm{grounding} &= R_\mathrm{soft} + R_\mathrm{hard},
\end{aligned}
\label{formula:grounding_reward}
\end{equation}
where  \({{g_\mathrm{last}}}\), \({{g_\mathrm{gt}}}\) represent the last grounding action and the ground truth label. The complementary nature of these reward signals motivates their integration: \(R_\mathrm{soft}\) offers discriminative power yet is relatively weak, whereas \(R_\mathrm{hard}\) delivers high-confidence guidance but is less discriminative. Their combination effectively offsets their respective limitations.

\noindent \textbf{Self-Confirmed Pseudo Reward.}
When dealing with large amounts of data without grounding labels, it is difficult to calculate \(R_\mathrm{grounding}\) to evaluate the quality of the grounding action. Distinguishing between better and worse grounding actions is non-trivial, as it has been observed that the model can sometimes produce correct answers while grounding video clips that do not contribute to answering the question, leading to unnecessary computational overhead. To address this issue, we propose a Self-Confirmed Pseudo Reward to approximate the effect of \(R_\mathrm{grounding}\) in encouraging high-quality grounding actions.

The core idea is to use the model itself as the judge. Specifically, we prompt the model to answer the question \(Q\) based solely on the last grounded video clip \(v_\mathrm{last}\). If the model can produce the correct answer using only this clip, then \(v_\mathrm{last}\) is considered more likely to be helpful for answering \(Q\) compared to when it fails to do so. 

In detail, for each trajectory containing grounding actions, let the last grounding action be \(g_\mathrm{last}\), the corresponding video clip be \(v_\mathrm{last}\) (which follows the fine-grained video representation defined in Section~\ref{preliminary} so as to accelerate with cache), and the ground-truth answer be \(a_\mathrm{gt}\). We prompt the policy model \(\mathcal{M}\) to generate an answer using only \(v_{last}\) and \(p'(Q)\) where \(p'\) is the prompt for this self-confirming instruction:
\begin{equation}
    a' = \mathcal{M}(v_\mathrm{last}, p'(Q)).
\end{equation}
Then, the \(R_{pseudo}\) is calculated through
\begin{equation}
    R_{\mathrm{pseudo}} = \begin{cases}
0, & \text{if the self-confirmed answer is correct},\\
-\gamma, & \text{otherwise},
\end{cases}
\label{formula:R_pseudo}
\end{equation}
where \(\gamma\) is a hyper-parameter, and the reward is implemented in the form of a penalty.

\noindent \textbf{Accuracy-Gated Mechanism.}
Since the correctness of the answer is of higher priority. Accordingly, the grounding-related reward is awarded only if the trajectory leads to a correct final answer to avoid the optimization conflict.

Above all, the reward can be calculated by
\begin{equation}
    R = \begin{cases}
        R_{\text{acc}} + R_{\text{format}}, & \text{(a)}\\[4pt]
        R_{\text{acc}} + R_{\text{format}} + \mathbb{I}(R_{\text{acc}}>0) \cdot R_{\text{grounding}}, & \text{(b)}\\[4pt]
        R_{\text{acc}} + R_{\text{format}} + \mathbb{I}(R_{\text{acc}}>0) \cdot R_{\text{pseudo}}, & \text{(c)}
    \end{cases}
\end{equation}
where (a), (b), and (c) correspond to trajectories without grounding actions, with grounding actions for labeled GQA data, and with grounding actions for general QA data, respectively. Algorithm~\ref{alg:twg-grpo} shows the reward calculation briefly.

\begin{algorithm}[t]
\caption{The process of rollout and reward calculation in our TwG-GRPO algorithm.}
\label{alg:twg-grpo}
\begin{algorithmic}[1]
\Require
The policy video LLM \( \mathcal{M}_{\theta} \); video \(V\); question \(Q\); ground truth answer \(a_\mathrm{gt}\); ground truth video grounding label \( g_\mathrm{gt} \) if there is; self-confirmed pseudo reward hyperparameter \( \gamma \); The maximum number of turns \(K\);
multi-grained video representation arguments  \( (F_\mathrm{coarse},f_\mathrm{coarse}) \) and \( (F_\mathrm{fine},f_\mathrm{fine}) \).
\Ensure
The trajectory \(\tau\); reward \(R\).

            \State \(v\) \( \leftarrow \)  VideoProcessor(\(V\), \(F_\mathrm{coarse}\), \(f_\mathrm{coarse}\))
            \State \(\tau\) \( \leftarrow \) [\(v\), \(p(Q)\) ]
            \For {\(k\  \text{\textbf{in}}\ 1,...,K\) }
                
                \State \(o_{k} \leftarrow \mathcal{M}_{\theta}(\tau) \) \Comment{Get the current output.}

                \If {\(o_k\) matches "grounding" format}
                    \State \(t_k\), \(g_k\) \(\leftarrow\) ParseGrounding(\(o_k\))
                    \State \(v_k\) \( \leftarrow \) VideoProcessor(VideoCut( \(V\), \(g_{k}\) ), \(F_\mathrm{fine}\), \(f_\mathrm{fine}\))
                    \State \(\tau\) \( \leftarrow \) \(\tau\) + [\(t_k\), \(g_k\), \(v_k\), \(p_k(Q)\)] \Comment{Conduct video grounding.}
                \ElsIf {\(o_k\) matches "answering" format}
                    \State \(t_k\), \(a\) \(\leftarrow\) ParseAnswering(\(o_k\))
                    \State \(\tau\) \( \leftarrow \) \(\tau\) + [\(t_k\), \(a\)] \Comment{Obtain the answer and stop.}
                    \State \textbf{Break}
                \Else
                    \State \(\tau\) \( \leftarrow \) \(\tau\) + [\(o_k\)] \Comment{Cannot parse it and stop.}
                    \State \textbf{Break}
                \EndIf
            \EndFor

        \State \(R\) \( \leftarrow \) \( R_\mathrm{acc} + R_\mathrm{format}\)
        \If {\(\tau\) has grounding actions}
            \State \( g_\mathrm{last} \) \( \leftarrow \) ExtractLastGrounding(\(\tau\))
            \If {the sample has label \(g_\mathrm{gt}\)}
                \State  \(R_\mathrm{grounding} \leftarrow\) GroundingReward(\(g_\mathrm{last}, g_\mathrm{gt}\)) \Comment{Eq. (\ref{formula:grounding_reward})}
                \State  \(R\) \( \leftarrow \) \( R +  \mathbb{I}(R_\mathrm{acc}>0) \cdot R_\mathrm{grounding} \)
            \Else
                \State  \(v_\mathrm{last}\)  \(\leftarrow\)  VideoProcessor(VideoCut( \(V\), \( g_\mathrm{last} \) ), \(F_\mathrm{fine}\),\(f_\mathrm{fine}\))
                \State \( a' \leftarrow  \mathcal{M}_{\theta} (v_\mathrm{last},p'(Q)) \)
                \State  \(R_\mathrm{pseudo} \leftarrow  \) PseudoReward( \(a',a_\mathrm{gt},\gamma\) ) \Comment{Eq.~(\ref{formula:R_pseudo})}
                \State  \(R\) \( \leftarrow \) \( R +  \mathbb{I}(R_\mathrm{acc}>0) \cdot R_\mathrm{pseudo} \)
            \EndIf
        \EndIf
\State \Return \(\tau\), \(R\)
\end{algorithmic}
\end{algorithm}

%% file: my_experiments.tex
\section{Experiments}

\begin{table*}[t]
\centering
\caption{Results of our Video-TwG and baselines on three mainstream long video benchmarks: Video-MME, LongVideoBench, MLVU. For MLVU, we evaluate the multiple choices questions. \(^*\) means our reproduced results. \(_{\textcolor{red}{\cdot \uparrow}}\) is the relative gain against the Qwen2.5-VL-7B. The best and second are marked with \textbf{bold} and \underline{underline}.}
\label{tab:main_results}
\small
\setlength{\tabcolsep}{5pt}
\renewcommand{\arraystretch}{1.15}

\begin{tabular}{lccccccccc c}
\toprule
\multirow{2}{*}{\textbf{Method}} &
\multicolumn{4}{c}{\textbf{Video-MME}} &
\multicolumn{5}{c}{\textbf{LongVideoBench}} &
\multirow{2}{*}{\textbf{MLVU-MC}} \\
\cmidrule(lr){2-5}\cmidrule(lr){6-10}
& short & medium & long & overall
& (8s,15s] & (15s,60s] & (180s,600s) & (900s,3600s] & overall
& \\

\midrule

Video-LLaVA-7B & 45.3 & 38.0 & 36.2 & 39.9 & - & - & - & - & 39.1 & 47.3 \\
ShareGPT4Video-8B & 48.3 & 36.3 & 35.0 & 39.9 & - & - & - & - & 39.7 & 46.4 \\
VideoChat2-7B & - & - &  33.2 & 39.5 & - & - & - & - & - & - \\
\midrule
LongVA-7B & 61.1 & 50.4 & 46.2 & 52.6 & - & - & - & - & 47.8 & 56.3 \\
Long-LLaVA-7B & 61.9 & 51.4 & 45.4 & 52.9 & - & - & - & - & - & - \\
LongLLaVA-9B  & 59.6 & 50.3 & 42.7 & 50.9 & - & - & - & - & 51.9 & - \\
Video-XL-7B & 64.0 & 53.2 & \underline{49.2} & 55.5 &  & - & - & - & 50.7 & - \\
\midrule

Video-R1-7B\(^*\) & \underline{69.3} & 53.6 & 49.1 & \underline{57.3} & - & - & - & - & \underline{55.2} & \underline{59.8} \\
Long-VILA-R1-7B\(^*\) & - & - & - & 34.6 & - & - & - & - & 35.5 & 36.6 \\
VideoChat-R1-7B\(^*\) & - & - & - & 54.4 & - & - & - & - & 40.8 & 57.0 \\

\midrule

VideoAgent & 53.3 & 49.7 & 37.8 & 46.9 & 54.6 & 55.2 & 45.1 & 43.5 & 47.1 & 52.5 \\
VideoTree & 55.5 & 49.2 & 39.3 & 48.0 & 61.0 & 57.5 & 48.4 & 44.6 & 49.7 & 51.6 \\

\midrule
Qwen2.5-VL-7B(LR) & 56.2 & 43.2 & 40.4 & 46.6 & 51.3 & 57.0 & 42.5 & 39.7 & 44.4 & 47.5 \\
\textbf{Video-TwG(LR)} & 63.7\(_{\textcolor{red}{7.5\uparrow}}\) & 49.4\(_{\textcolor{red}{6.2\uparrow}}\)& 47.7\(_{\textcolor{red}{7.3\uparrow}}\)& 53.6\(_{\textcolor{red}{7.0\uparrow}}\)& 59.8\(_{\textcolor{red}{8.5\uparrow}}\)& 62.8\(_{\textcolor{red}{5.8\uparrow}}\)& 47.3\(_{\textcolor{red}{4.8\uparrow}}\)& 44.1\(_{\textcolor{red}{4.4\uparrow}}\)& 49.7\(_{\textcolor{red}{5.3\uparrow}}\)& 54.6\(_{\textcolor{red}{7.1\uparrow}}\)\\
Qwen2.5-VL-7B(HR) & 71.2 & \underline{53.8} & 46.7 & 57.2 & \underline{63.5} & \underline{69.8} & \underline{49.5} & \underline{45.6} & {52.4} & 55.3 \\
\textbf{Video-TwG(HR)} & \textbf{71.9}\(_{\textcolor{red}{0.7\uparrow}}\)& \textbf{57.1}\(_{\textcolor{red}{3.3\uparrow}}\)& \textbf{50.0}\(_{\textcolor{red}{3.3\uparrow}}\)& \textbf{59.7}\(_{\textcolor{red}{2.5\uparrow}}\)& \textbf{73.5}\(_{\textcolor{red}{10.0\uparrow}}\)& \textbf{72.7}\(_{\textcolor{red}{2.9\uparrow}}\)& \textbf{54.1}\(_{\textcolor{red}{4.6\uparrow}}\)& \textbf{47.2}\(_{\textcolor{red}{1.6\uparrow}}\)& \textbf{56.3}\(_{\textcolor{red}{3.9\uparrow}}\)& \textbf{60.3}\(_{\textcolor{red}{5.0\uparrow}}\)\\
\bottomrule
\end{tabular}
\end{table*}

\subsection{Implementations}

We employ Qwen2.5-VL-7B~\cite{bai2025qwen2} as the base video LLM. For efficient training, we adopt LoRA~\cite{hu2022lora} with a rank of 128 and an alpha of 256. The learning rate is set to \(1 \times 10^{-6}\). During GRPO training, a batch size of 32 with a group size of 8 is used. The KL constraint coefficient \(\beta\) is set to 0.005, and the coefficient \(\gamma\) in the Self-Confirmed Pseudo Reward is set to 0.1. For RL sampling, the temperature, top-p, top-k, and repetition penalty are configured as 1.0, 0.9, 50, and 1.0, respectively, with a maximum turn count \(K = 3\). 
For coarse-grained video representation during training, we sample \(F_{\mathrm{coarse}} = 64\) frames, allocating up to \(f_{\mathrm{coarse}} = 16\) tokens per frame. For fine-grained representation, we sample \(F_{\mathrm{fine}} = 16\) frames with a maximum of \(f_{\mathrm{fine}} = 64\) tokens per frame.
During inference, the frame counts remain fixed as above, while the maximum number of tokens per frame is adjusted via input video resolution to align to other baselines on the input length. We evaluate our Video-TwG under two resolution settings: (i) {LR (Low Resolution):} Matches the training configuration, allowing at most 16 tokens per frame for coarse-grained and 64 tokens per frame for fine-grained representations. (ii) {HR (High Resolution):} Allows at most 128 tokens per frame for coarse-grained and 512 tokens per frame for fine-grained representations.
In evaluation, temperature is set to 0 and we set a maximum turn count \(K = 3\) as well. If a prediction cannot be automatically parsed, we revert to the training sampling hyperparameters and retry up to three times. For a fair comparison, both our base model Qwen2.5-VL-7B and other evaluated models that support reasoning are assessed in reasoning mode.
Our training and evaluation framework is built upon ms-swift~\cite{zhao2024swiftascalablelightweightinfrastructure} and VLMEvalKit~\cite{duan2024vlmevalkit}. We leverage vLLM~\cite{kwon2023efficient} to accelerate the sampling process. Our training can be implemented on 2 H800 80G GPUs with early-stop strategy.

\subsection{Baselines}

The baselines used in our experiments are categorized into four groups:
(i) {General video LLMs}, including Video-LLaVA~\cite{lin2024video}, 
Share\-GPT4Video~\cite{chen2024sharegpt4video}, and VideoChat2~\cite{li2024mvbench};
(ii) {Long-context video LLMs}, such as LongVA~\cite{zhang2024long}, 
Long-LLaVA~\cite{long-llava-qwen2-7b-2024}, 
LongLLaVA~\cite{wang2024longllava}, and Video-XL~\cite{shu2025video};
(iii) {Video agentic systems for long video understanding}, represented by 
VideoAgent~\cite{wang2024videoagent} and VideoTree~\cite{wang2025videotree}; and
(iv) {Video reasoning models}, including Video-R1~\cite{feng2025video}, 
VideoChat-R1~\cite{li2025videochat}, and Long-VILA-R1~\cite{chen2025scaling}.

\subsection{Benchmarks}

We evaluate our method on multiple-choice questions across three mainstream benchmarks for long video understanding:
Video-MME~\cite{fu2025video}, LongVideoBench~\cite{wu2024longvideobench}, and MLVU~\cite{zhou2025mlvu}.
Video-MME consists of 2,700 manually annotated multiple-choice questions derived from 900 videos.
The videos are divided into three groups based on duration: short (0-2 minutes), medium (4-15 minutes), and long (30-60 minutes).
LongVideoBench contains 3,763 human-labeled questions spanning 6,678 videos, with an average duration of 472 seconds.
The videos are categorized into four duration ranges: 8-15 seconds, 15-60 seconds, 180-600 seconds, and 900-3,600 seconds.
We evaluate models on its validation split, which includes 1,337 multiple-choice questions.
MLVU features videos ranging from 3 to 120 minutes, with an average duration of 15 minutes.
It provides both Dev and Test splits, comprising multiple-choice and open-ended questions.
In this work, we focus on the multiple-choice questions in the Dev split, which contains 2,174 questions covering seven distinct question types.
For all benchmarks, subtitles are not used during evaluation.

\subsection{Main Results}

Table~\ref{tab:main_results} summarizes the performance of Video-TwG and a wide range of baselines on three representative benchmarks for long video understanding.
Overall, Video-TwG achieves consistently strong performance across datasets with diverse video durations, outperforming general video LLMs, long-context video LLMs, as well as recent reasoning and agent-based approaches.

Compared with general-purpose video LLMs, Video-TwG yields substantial improvements on all benchmarks.
On Video-MME, Video-TwG (HR) achieves an overall accuracy of 59.7\%, significantly outperforming general video LLMs that typically obtain around 40\%.
It also attains 50.0\% accuracy on long videos, exceeding prior methods by a large margin, indicating that models primarily designed for short or medium-length videos struggle to generalize to long-duration scenarios.
Video-TwG also consistently outperforms long-context models including LongVA, Long-LLaVA, LongLLaVA, and Video-XL.
For example, on Video-MME, Video-TwG (HR) exceeds Video-XL by over 4 points in overall accuracy, suggesting that effective long-range reasoning is crucial beyond simply increasing the input context length.
Recent video reasoning models, such as Video-R1, enhance long video understanding by introducing explicit reasoning mechanisms.
While effective, these methods rely on reasoning under fixed video context which is not enough for very long temporal contexts, since the model is observed to make up facts confidently when the video information is not enough, which is shown in the case study.
In contrast, Video-TwG achieves stronger performance via grounding and zooming in, indicating more effective long-range temporal modeling.
Agent-based systems, including VideoAgent and VideoTree, address long video understanding through explicit planning or hierarchical decomposition.
Despite their improvements over earlier baselines, they depend on complex multi-stage pipelines that are prone to error accumulation and heavy computations, like captioning.
Video-TwG consistently outperforms VideoTree by 11.7, 6.6 and 8.7 points, and VideoAgent by 12.8, 9.2 and 7.8 points on these benchmarks, respectively, while maintaining a simpler inference paradigm.

Furthermore, under both low-resolution (LR) and high-resolution (HR) settings, Video-TwG consistently outperforms its backbone Qwen-2.5-VL-7B across all benchmarks.
The improvements are more pronounced in the low-resolution setting, while models trained under low-resolution still generalize effectively to high-resolution inference.
This indicates that the gains primarily stem from the proposed training paradigm, making Video-TwG particularly effective for real-world and resource-constrained scenarios.

\subsection{Ablation Study}

We further conduct ablation study to explore the contributions of the core components of our Video-TwG method on the Video-MME benchmark. By default, we adopt the low-resolution setup for its friendly computational costs.

\begin{table}[t]
\centering
\caption{Ablation experiments of different variants of the base Qwen2.5-VL-7B on Video-MME.}
\label{tab:abla_twg_fashion}
\small
\setlength{\tabcolsep}{5pt}
\renewcommand{\arraystretch}{1.15}

\begin{tabular}{lcccc}
\toprule
\multirow{2}{*}{\textbf{Method}} &
\multicolumn{4}{c}{\textbf{Video-MME}} \\
\cmidrule(lr){2-5}
& short & medium & long & overall \\
\midrule
Base$^{1}$ & 56.2 & 43.2 & 40.4 & 46.6 \\
Base-TwG-CoT$^{1}$ & 56.9 & 43.2 & 42.1 & 47.4 \\
Base-Think$^{2}$ & 59.2 & 46.4 & 43.2 & 49.6 \\
Base-MultiGrained$^{4}$ & 63.1 & 45.9 & 42.7 & 50.6 \\
Base-Think-MultiGrained $^{5}$ & 63.0 & 46.9 & 45.3 & 51.7 \\
\textbf{Video-TwG} & 63.7 & 49.4 & 47.7 & 53.6  \\
\bottomrule
\end{tabular}

\vspace{2pt}
\footnotesize
\begin{minipage}[t]{\linewidth}
\raggedright
$^{1}$ {Base}: Qwen2.5-VL-7B. \\
$^{2}$ {Base-TwG-CoT}: Base is prompted to perform \textit{think-with-grounding} using the same prompts as Video-TwG. \\
$^{3}$ {Base-Think}: Base is trained on the TwG-51K dataset to conduct text-only reasoning without grounding. \\
$^{4}$ {Base-MultiGrained}: Base receives both coarse- and fine-grained representations of the entire video as input.\\
$^{5}$ {Base-Think-MultiGrained}: Base-Think receives both coarse- and fine-grained representations of the entire video as input.
\end{minipage}

\end{table}

\noindent \textbf{Analysis on the \textit{think-with-grounding} paradigm.}
In Tab.~\ref{tab:abla_twg_fashion}, we show several experiments on the base Qwen2.5-VL-7B model for analysis on the \textit{think-with-grounding} paradigm itself. These experiments are conducted to answer these questions. \textit{(i) Can the model benefit from \textit{think-with-grounding}?} We directly prompt the base model to conduct \textit{think-with-grounding} with the same prompts as applied in Video-TwG~(Base-TwG-CoT in Tab.~\ref{tab:abla_twg_fashion}).
We can see that in the short and medium splits, the improvement is trivial but it has a 1.7 performance gain in the long video split, showing that the base model has the potential to conduct \textit{think-with-grounding} reasoning to improve the performance, especially in long videos. However, there is still a significant gap between Base-TwG-CoT and our Video-TwG, proving the effectiveness of our method to further enhance \textit{think-with-grounding} ability.
\textit{(ii) Can \textit{think-with-grounding} go beyond traditional text-only reasoning?} We train the base model on the same TwG-51K dataset to make the model conduct text-only reasoning without grounding actions with the GRPO algorithm, whose results are shown in "Base-Think" in Tab.~\ref{tab:abla_twg_fashion}.
Although it achieves improvement compared to the base, there is still a significant gap compared to our method. It is 4.5 lower than our method in long videos.
Our method allows the model to dynamically load fine-grained video to context, which improves the model performance, while text-only RL only helps the model to better grasp the clues in the given fixed multimodal input.
\textit{(iii) Since \textit{think-with-grounding} dynamically visits more video input, can such paradigm go beyond directly scaling the video input length?} In the Video-MME benchmark, our Video-TwG is observed to conduct 0.65 grounding actions on average per sample. To answer this question, we increase the video input length for the base model, as shown in Base-MultiGrained and Base-Think-MultiGrained in Tab.~\ref{tab:abla_twg_fashion}, and we feed the Base and Base-Think with both the coarse- and fine-grained representation of the entire video, which is equivalent to conducting 1 grounding action per sample on the video context length. It shows that such vanilla scaling nearly catches up with our Video-TwG in short videos but there is a significant gap of 5.0(Base-MultiGrained) and 2.4(Base-Think-MultiGrained) in long videos, showing the necessity of intelligent selection of video clip, especially in long videos.

\begin{figure}[h]
  
    \centering
    \includegraphics[width=\linewidth]{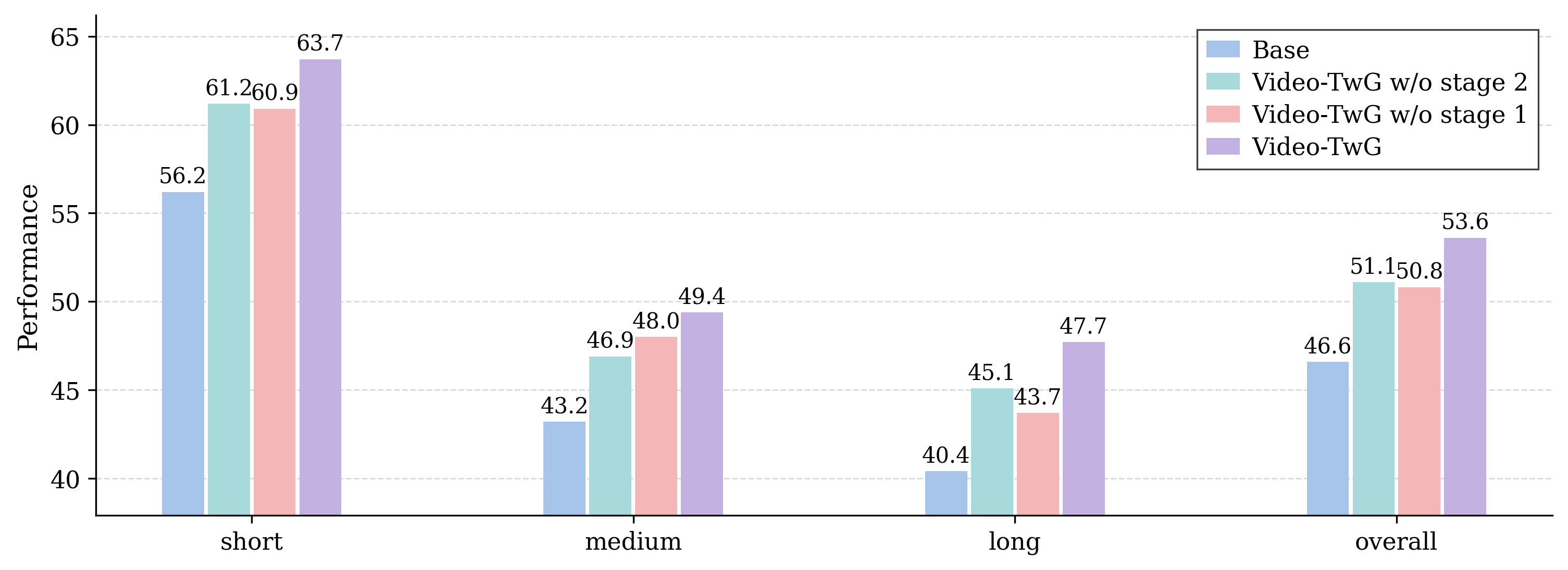}
    \caption{Ablation results on Video-MME of our models with different training stages.}
    \Description{}
    \label{fig:two-stage-abla}
\end{figure}

\noindent \textbf{Analysis on our two-stage curriculum training strategy.} We compare our Video-TwG with two variants, when it's trained with stage 1 or 2 only. The results are shown in Fig.~\ref{fig:two-stage-abla}, which shows that the performance drops significantly when either of training stages is skipped. It's noted that if skipping the stage 1, the model's grounding action vanishes quickly in the early stage of the training during stage 2, indicating the necessity of stage 1 as a good cold-start, and the the model's performance in long videos is 1.4 inferior to when it is trained in stage 1. When equipped with stage 1 and stage 2, the model gradually grasp the \textit{think-with-grounding} ability and achieves the best performance.

\begin{figure}[h]
  \centering

  \begin{subfigure}[b]{0.49\linewidth}
    \centering
    \includegraphics[width=\linewidth]{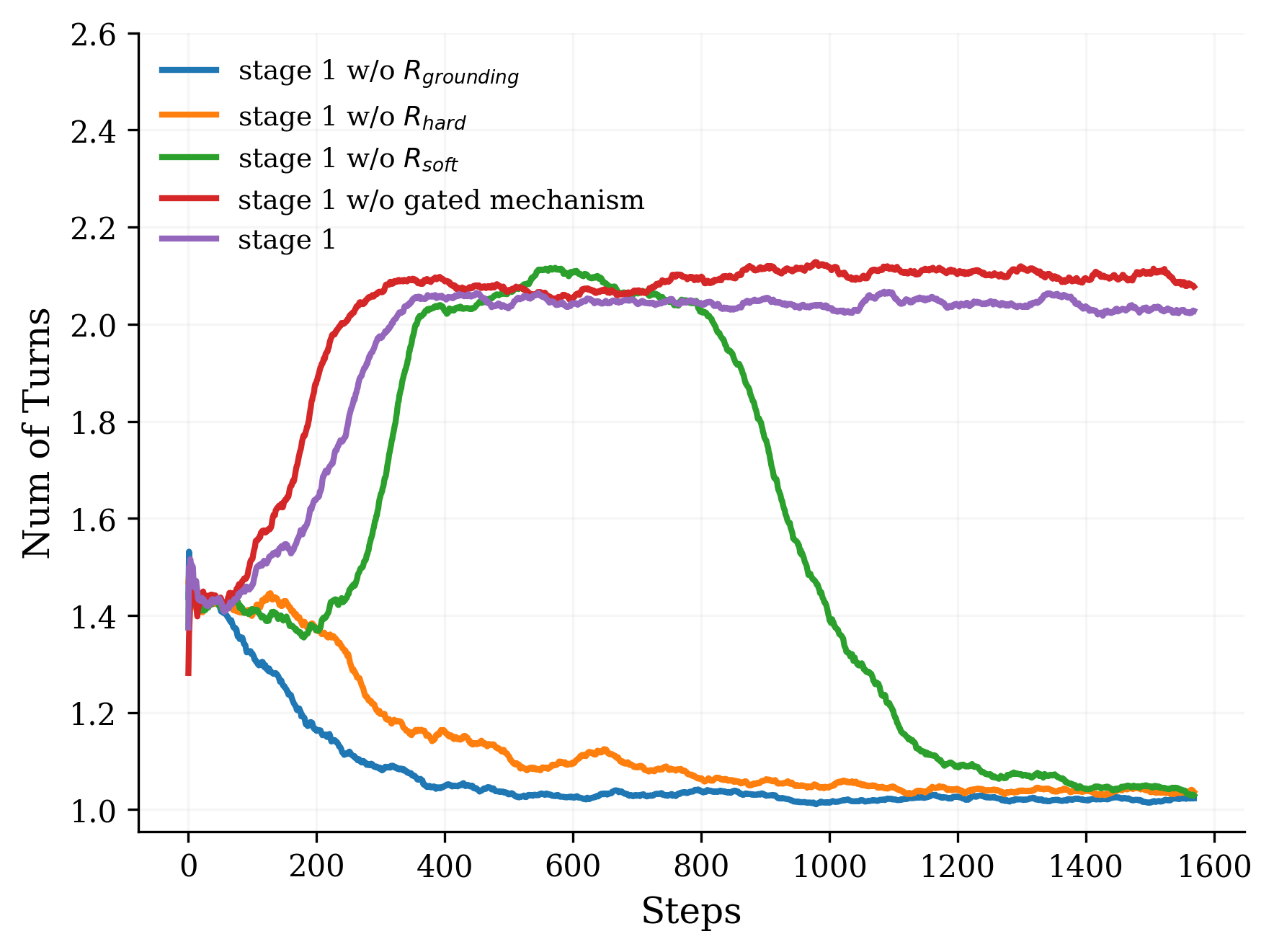}
    \caption{The number of turns}
    \Description{}
    \label{fig:R-grounding_abla_num_turns}
  \end{subfigure}
  \hfill
  \begin{subfigure}[b]{0.49\linewidth}
    \centering
    \includegraphics[width=\linewidth]{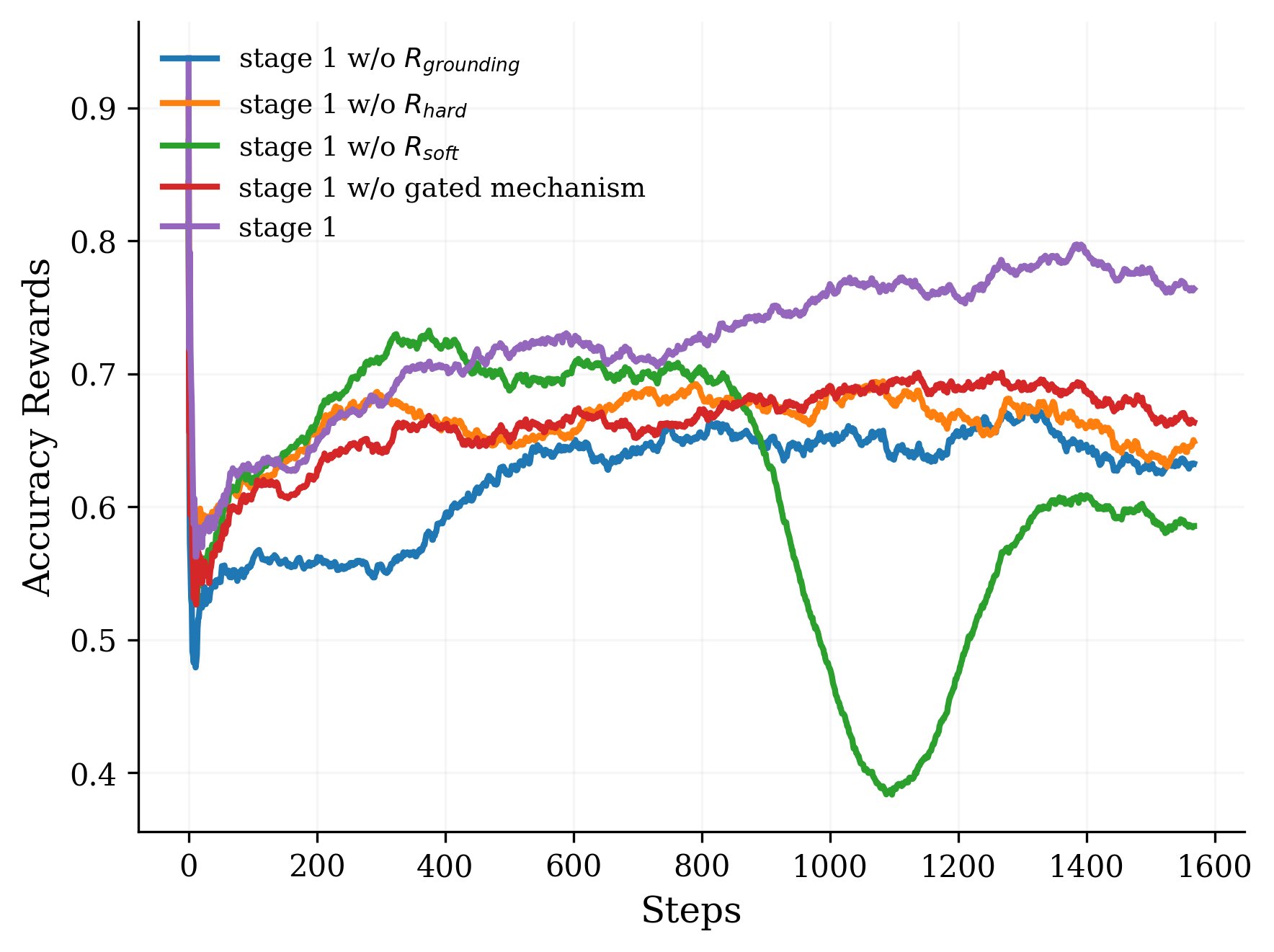}
    \caption{The accuracy reward}
    \Description{}
    \label{fig:R-grounding_abla_acc}
  \end{subfigure}
  \caption{Training curves of several variants of our Video-TwG in stage 1. The curves are smoothed for aesthetics. It's the same below.}
  \label{fig:stage_1_R_grounding_abla}
  \Description{}
\end{figure}

\noindent \textbf{Analysis around the grounding reward in stage 1 training.}
We explore the roles of several core factors in stage 1 training, the accuracy-gated mechanism, the grounding reward \(R_\mathrm{grounding}\), the soft grounding reward \(R_\mathrm{soft}\), and the hard grounding reward \(R_\mathrm{hard}\). Fig.~\ref{fig:stage_1_R_grounding_abla} shows the curves of the number of turns and the accuracy reward during the stage 1 training. We can see that without \(R_\mathrm{grounding}\) or without \(R_\mathrm{hard}\) in \(R_\mathrm{grounding}\), the grounding action vanishes along with the training process, showing that the signal provided by \(R_\mathrm{soft}\) is too weak to boost the model to learning \textit{think-with-grounding}. Only with \(R_\mathrm{hard}\) in \(R_\mathrm{grounding}\), the training is unstable where the model collapses at around 800 steps when the model quickly forgets the grounding action. Compared to \(R_\mathrm{soft}\), \(R_\mathrm{hard}\) fails to provide diversified and differentiated rewards in a group, which makes the training process less stable compared to \(R_\mathrm{soft}+R_\mathrm{hard}\). Without the accuracy gated mechanism, the model learns slightly quickly to think with grounding, but however, it achieves much lower accuracy rewards than our standard model in stage 1. The grounding and accuracy targets may conflict on some samples, thereby affecting the performance. We conduct quantitative evaluation for model without accuracy-gated mechanism in stage 1, as shown in Tab.~\ref{tab:stage-1-abla-gated}, which shows it is 0.5 and 1.5 lower than the standard model in the long and overall splits.

\begin{table}[t]
\centering
\caption{The ablation results of the accuracy-gated mechanism in stage 1 training.}
\label{tab:stage-1-abla-gated}
\small
\setlength{\tabcolsep}{5pt}
\renewcommand{\arraystretch}{1.15}

\begin{tabular}{lcccc}
\toprule
\multirow{2}{*}{\textbf{Method}} &
\multicolumn{4}{c}{\textbf{Video-MME}} \\
\cmidrule(lr){2-5}
& short & medium & long & overall \\
\midrule
stage 1 w/o gated mechanism & 60.2 & 48.1 & 43.6 & 50.6 \\
\textbf{stage 1} & 61.2 & 46.9 & 45.1 & 51.1 \\

\bottomrule
\end{tabular}
\label{tab:abla}
\end{table}

\begin{figure}[h]
  \centering

  \begin{subfigure}[b]{0.49\linewidth}
    \centering
    \includegraphics[width=\linewidth]{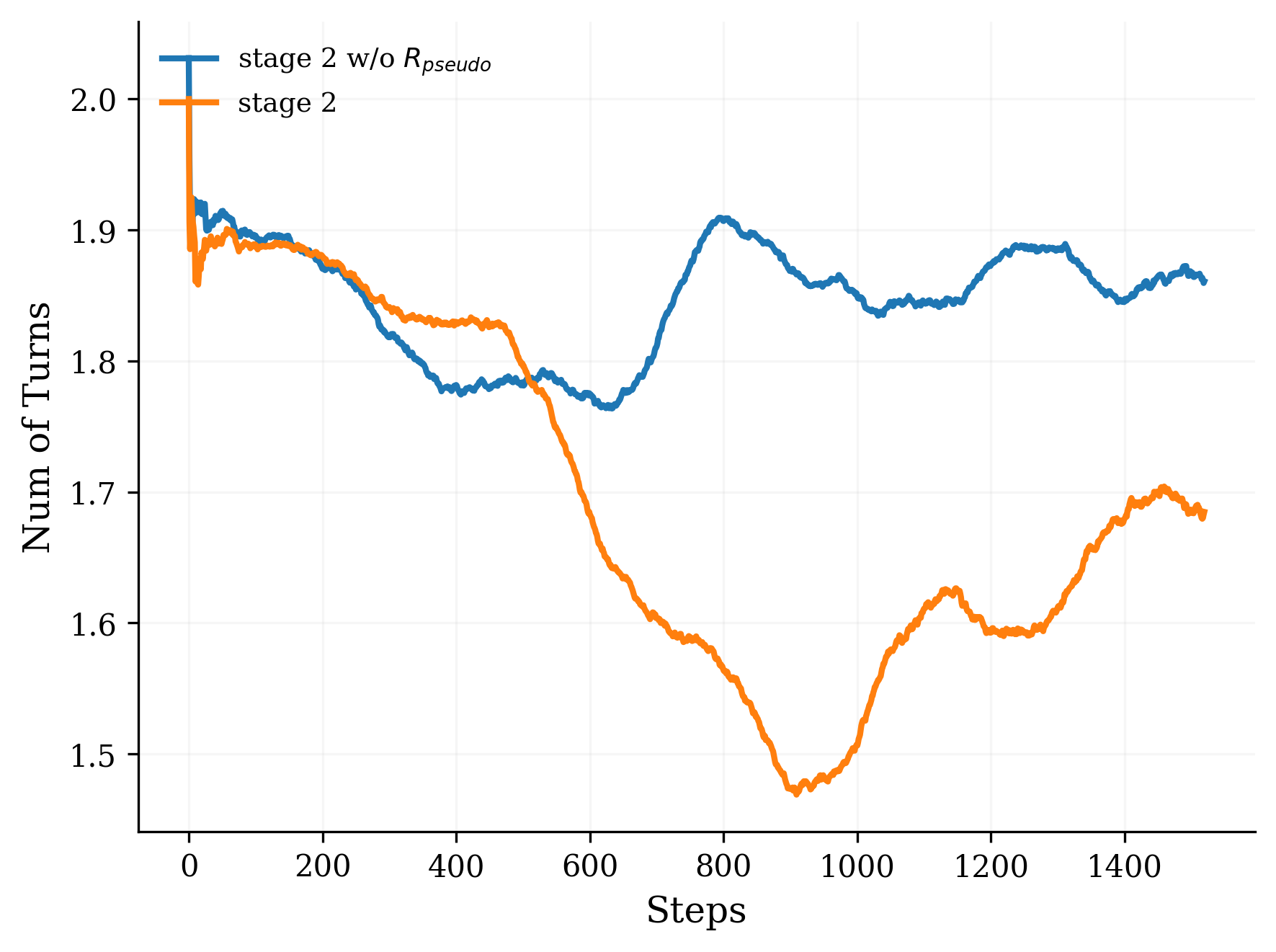}
    \caption{The number of turns.}
    \Description{}
    \label{fig:R_conf_abla_num_turns}
  \end{subfigure}
  \hfill
  \begin{subfigure}[b]{0.49\linewidth}
    \centering
    \includegraphics[width=\linewidth]{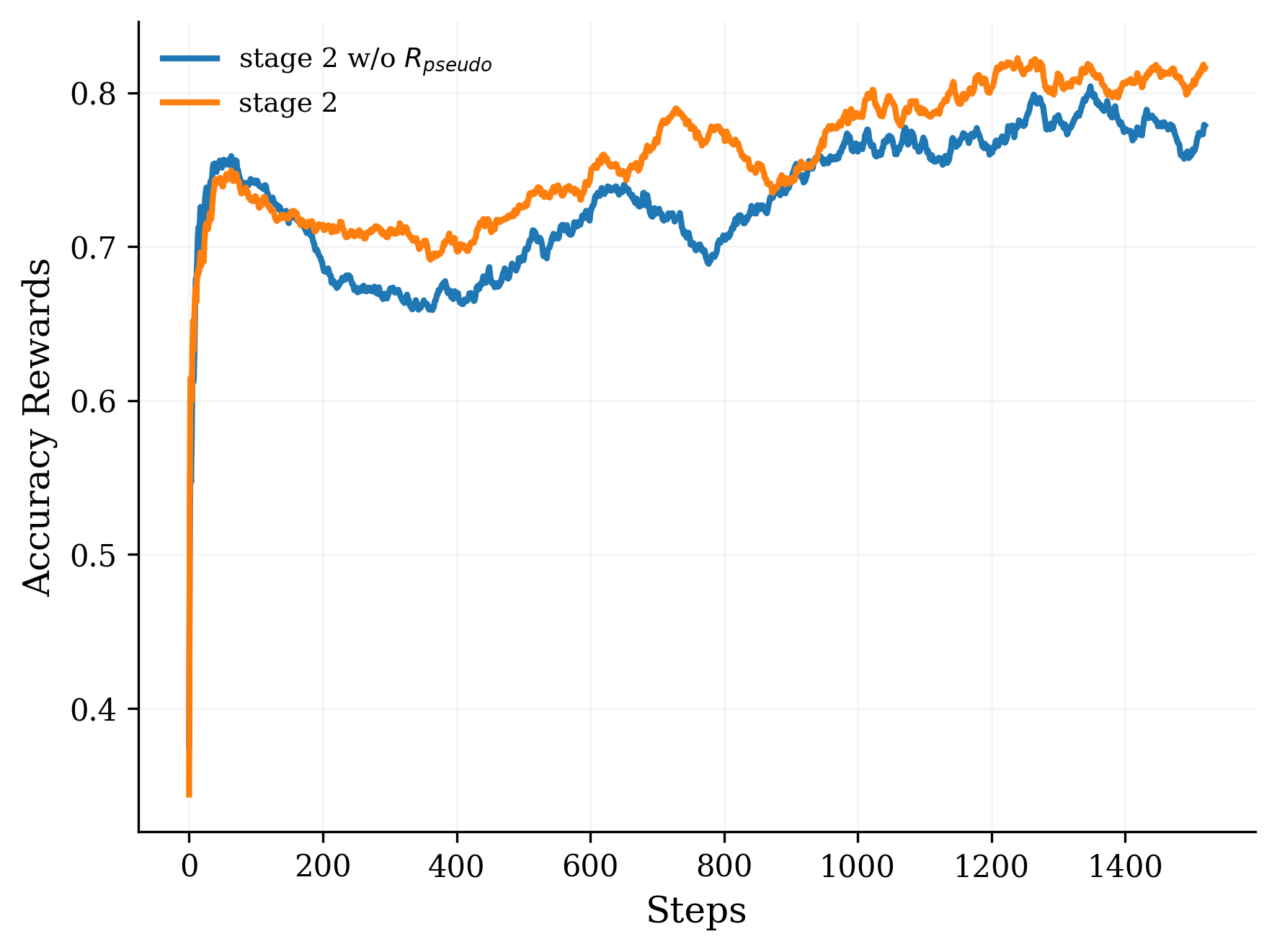}
    \caption{The accuracy rewards.}
    \Description{}
    \label{fig:R_conf_abla_acc}
  \end{subfigure}
  \caption{Training curves of several variants of our Video-TwG in stage 2.}
  \label{fig:stage_2_R_pseudo_abla}
  \Description{}
\end{figure}

\begin{table}[t]
\centering
\caption{The ablation results of the self-confirmed pseudo reward in stage 2 training.}
\label{tab:stage-2-pseudo-abla}
\small
\setlength{\tabcolsep}{5pt}
\renewcommand{\arraystretch}{1.15}

\begin{tabular}{lcccc}
\toprule
\multirow{2}{*}{\textbf{Method}} &
\multicolumn{4}{c}{\textbf{Video-MME}} \\
\cmidrule(lr){2-5}
& short & medium & long & overall \\
\midrule
Video-TwG w/o \(R_\mathrm{pseudo}\) & 63.4 & 48.0 & 47.2 & 52.9 \\
\textbf{Video-TwG} & 63.7 & 49.4 & 47.7 & 53.6  \\
\bottomrule
\end{tabular}
\end{table}

\begin{figure*}[h]
  \centering
  \includegraphics[width=\linewidth]{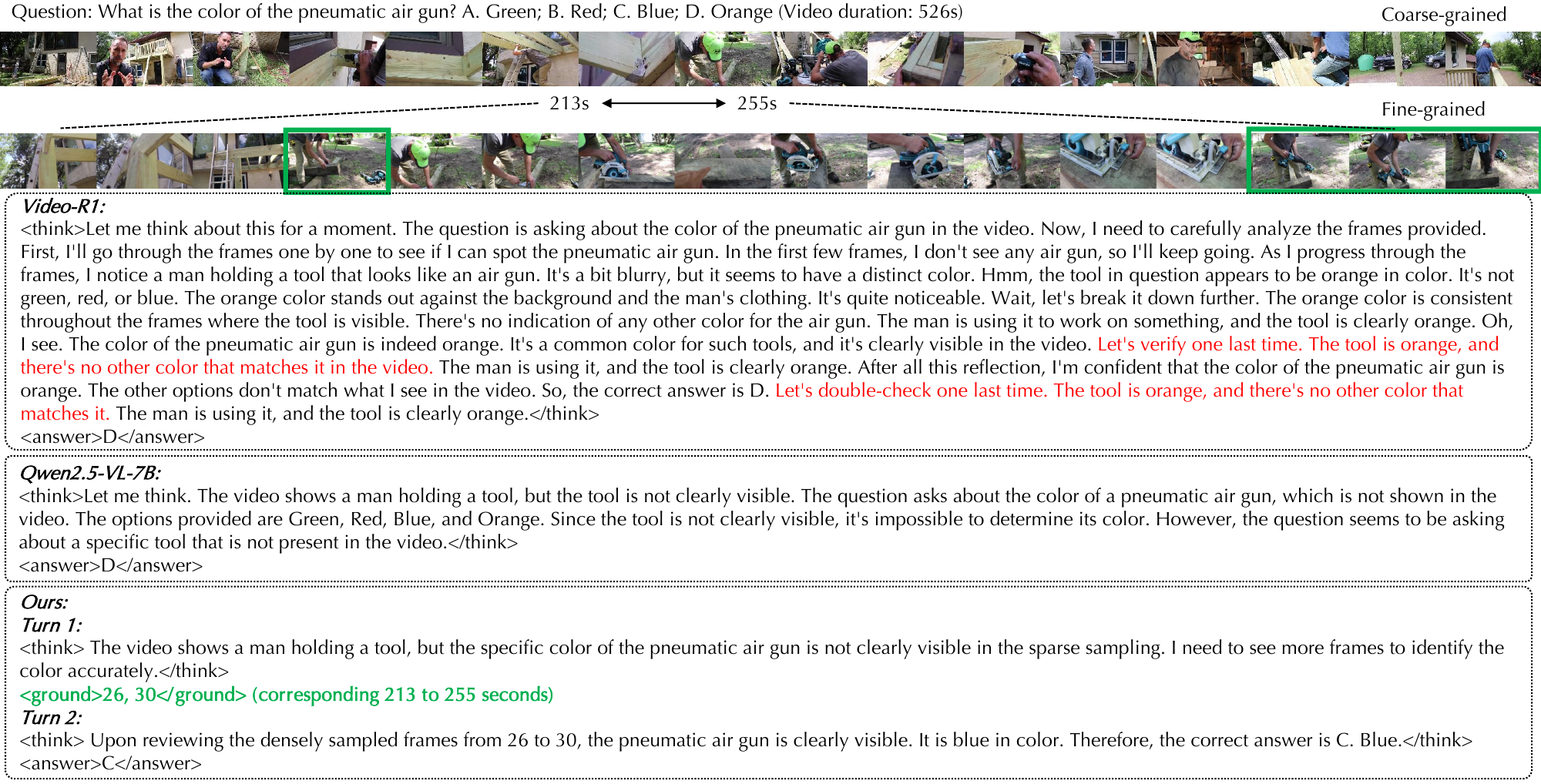}
  \caption{The case study on Video-R1, Qwen2.5-VL-7B, Our Video-TwG. The location of the crucial video clues for this question is marked with \textcolor{ForestGreen}{green} boxes.}
  \label{fig:case-study}
  \Description{}
\end{figure*}

\noindent \textbf{Analysis on the Self-Confirmed Pseudo Reward \(R_\mathrm{pseudo}\) in stage 2 training.}
Fig.~\ref{fig:stage_2_R_pseudo_abla} illustrates the training dynamics of the number of turns and accuracy reward during stage~2. Both settings exhibit a similar trend: the number of grounding actions first decreases and then increases. This suggests that grounding is not always necessary for all questions, the model learns to answer questions with fewer grounding actions in the early stage of training, leading to a reduction in turns. As training progresses, the model realizes that additional grounding can further improve accuracy, and the number of turns begins to rise. Correspondingly, accuracy rewards start to increase more steadily after the turning points (around 700 steps for "stage 2 w/o \(R_\mathrm{pseudo}\) " and 900 steps for "stage 2").
With the introduction of the Self-Confirmed Pseudo Reward \(R_\mathrm{pseudo}\), the model learns to invoke grounding actions more adaptively, using fewer grounding steps on average while achieving higher accuracy rewards throughout training. As shown in Tab.~\ref{tab:stage-2-pseudo-abla}, removing \(R_\mathrm{pseudo}\) leads to a slight performance drop, demonstrating its effectiveness in reducing redundant grounding actions without sacrificing overall performance.

We further analyze the number of samples containing grounding actions for Video-TwG w/o \(R_\mathrm{pseudo}\) and Video-TwG in Tab.~\ref{tab:stage-2-abla-grounding-times}. Video-TwG reduces the number of such samples by 15.7\%, approximately 300, indicating more selective grounding behavior. To assess the grounding quality, we evaluate QA performance when only the last grounded video clip is provided as input as an approximation to the grounding quality. For objectivity, we employ InternVL-3.5-38B~\cite{wang2025internvl3} as an external evaluator, sampling 32 frames with 256 tokens per frame. Results in Tab.~\ref{tab:grounding-quality-eval} show that grounding clips produced by standard Video-TwG yield better QA performance, with gains of 2.1 and 1.6 on the long and overall splits, respectively.

Overall, these results indicate that \(R_\mathrm{pseudo}\) encourages more effective and cautious grounding strategies, particularly for reducing useless exploration in long videos. The model learns to avoid unnecessary grounding actions that do not contribute to answering the question, leading to both higher efficiency and better grounding quality.

\begin{table}[t]
\centering
\caption{The number of samples that has grounding action on Video-MME benchmark.}
\label{tab:stage-2-abla-grounding-times}
\small
\setlength{\tabcolsep}{5pt}
\renewcommand{\arraystretch}{1.15}

\begin{tabular}{lcccc}
\toprule
\multirow{2}{*}{\textbf{Method}} &
\multicolumn{4}{c}{\textbf{Video-MME}} \\
\cmidrule(lr){2-5}
& short & medium & long & overall \\
\midrule
Video-TwG w/o \(R_\mathrm{pseudo}\) & 682 & 687 & 725 & 2094 \\
Video-TwG & 615 & 584 & 566 & 1765 \\

\bottomrule
\end{tabular}
\end{table}

\begin{table}[t]
\centering
\caption{The evaluation results on the grounded video clip on Video-MME by InternVL-3.5-38B for our Video-TwG w/ and w/o \(R_\mathrm{pseudo}\).}
\label{tab:grounding-quality-eval}
\small
\setlength{\tabcolsep}{5pt}
\renewcommand{\arraystretch}{1.15}

\begin{tabular}{lcccc}
\toprule
\multirow{2}{*}{\textbf{Method}} &
\multicolumn{4}{c}{\textbf{Video-MME}} \\
\cmidrule(lr){2-5}
& short & medium & long & overall \\
\midrule
Video-TwG w/o \(R_\mathrm{pseudo}\) & 65.5 & 55.3 & 51.4 & 57.3 \\
Video-TwG & 68.1 & 54.5 & 53.5 & 58.9 \\

\bottomrule
\end{tabular}
\end{table}

\subsection{Case Study}

In this section, we present a qualitative case study comparing the responses of Video-R1, Qwen2.5-VL-7B, and our method, which is shown in Fig.~\ref{fig:case-study}.
Given a fixed and sparse video context from a long video, Video-R1 produces substantially longer reasoning traces than Qwen2.5-VL-7B; however, its reasoning process is chaotic and leads to hallucinated conclusions, confidently identifying the tool as orange despite being incorrect.
The base model Qwen2.5-VL-7B recognizes that key visual clues are missing in the initial coarse-grained video input, which ultimately results in an incorrect answer.
In contrast, our method explicitly expresses uncertainty in its initial prediction and performs a grounding action in the first turn.
After receiving fine-grained visual input from the grounded video clip, it successfully identifies the correct answer.

%% file: my_conclude.tex
\section{Conclusion}

In this work, we introduce {Video-TwG}, a curriculum reinforced framework that enables video LLMs to think with grounding in a multi-turn reasoning process, zooming into question-relevant clips only when necessary. To support scalable training, we construct the {TwG-51K} dataset and propose a {two-stage reinforced curriculum strategy} that cold-starts on short-video GQA with grounding labels and then scales to the full dataset to improve generalization. Moreover, to handle complicated \textit{think-with-grounding} reasoning for various kinds of data, we develop a {TwG-GRPO} algorithm which features the fine-grained grounding reward, self-confirmed pseudo reward, and accuracy-gated mechanism. 
Extensive experiments on Video-MME, LongVideoBench, and MLVU demonstrate consistent advantages over strong baselines. Our results validate the effectiveness of our two-stage reinforced curriculum strategy: training with low-resolution inputs consistently enhances the base model’s \textit{think-with-grounding} ability across different inference resolutions, while {TwG-GRPO} better leverages diverse unlabeled data to improve grounding quality and reduce redundant groundings without sacrificing QA performance.
In future work, we plan to go beyond trajectory-level optimization by developing fine-grained process reward modeling and single-step training under more complex constraints, aiming to build more powerful and efficient long video reasoning models.